\pdfoutput=1

\documentclass[11pt]{article}

\usepackage{acl}

\usepackage{times}
\usepackage{latexsym}

\usepackage[T1]{fontenc}

\usepackage[utf8]{inputenc}

\usepackage{microtype}

\usepackage{graphicx}
\usepackage[russian,english]{babel}
\usepackage{color}

\title{WikiOmnia: filtration and evaluation of the generated QA corpus on the whole Russian Wikipedia}

\author{Dina Pisarevskaya \\
  Independent Researcher \\
  \texttt{dinabpr@gmail.com} \\\And
  Tatiana Shavrina \\
  Artificial Intelligence Research Institute (AIRI) \\
  SberDevices \\
  \texttt{rybolos@gmail.com} \\}

\begin{document}
\maketitle
\begin{abstract}
The General QA field has been developing the methodology referencing the Stanford Question answering dataset (SQuAD) as the significant benchmark. 
Compiling factual questions datasets requires manual annotations, limiting the training data's potential size. We present the WikiOmnia dataset, a new publicly available set of QA pairs and corresponding Russian Wikipedia article summary sections, composed with a fully automated generation and filtration pipeline. To ensure high quality of generated QA pairs, diverse manual and automated evaluation techniques were applied. The WikiOmnia pipeline is available open-source and is also tested for creating SQuAD-formatted QA on other domains, like news texts, fiction, and social media.
The resulting dataset includes two parts: raw data on the whole Russian Wikipedia (7,930,873 QA pairs with paragraphs for ruGPT-3 XL and 7,991,040 QA pairs with paragraphs for ruT5-large) and cleaned data with strict automatic verification (over 160,000 QA pairs with paragraphs for ruGPT-3 XL and over 3,400,000 QA pairs with paragraphs for ruT5-large).
\end{abstract}
\section{Introduction}

Generative abilities of large and high-performing pre-trained language models (LMs) are widely investigated now, and special interest is aroused around generating datasets in a fully unsupervised way~\citep{schick-schutze-2021-generating}. 
Question answering (QA) datasets can be easily adjusted to the generation pipeline formats and become a source for training generative reading comprehension  systems~\citep{brown2020language,wei2021finetuned}, dialogue systems~\citep{nehring-etal-2021-combining}, various tasks in the field of information retrieval for various languages~\citep{pisarevskaya2021}.

In this work, we present WikiOmnia - the largest QA dataset for Russian, obtained in a fully-automated way. The dataset contains QA pairs for every article of Russian Wikipedia \footnote{as of March 2021}, based on the summary sections. WikiOmnia consists of 2 parts: 
\begin{enumerate}
    \item the voluminous, automatically generated part: 15,9 million triplets consisting of the original article summary, a corresponding generated question and a generated answer; 
    \item the filtered part: the subsample of 3,5 million triplets, fully verified with automatic means.
\end{enumerate}
Apart from the data, we present a fully-automated pipeline for SQuAD-like data generation for Russian, based on \textit{generative part} represented by the ruGPT-3 XL\footnote{\url{https://huggingface.co/sberbank-ai/rugpt3xl}} and ruT5-large \footnote{\url{https://huggingface.co/sberbank-ai/ruT5-large}} models, and \textit{filtering part} that includes Russian BERT\footnote{\url{http://docs.deeppavlov.ai/en/master/features/models/squad.html}} baseline and rich heuristic approach.
All stated models were fine-tuned on SberQuAD~\citep{efimov2020sberquad} that is based on the methodology of the original English SQuAD~\citep{rajpurkar2016squad}. 
The whole automated and unsupervised generation and filtration pipeline was also tested for creating SQuAD-formatted QA on other domains: news texts, customer reviews, fiction, and social media. 
QA datasets generated with ruGPT3XL and ruT5 will be available   \href{https://huggingface.co/datasets/RussianNLP/wikiomnia}{on HuggingFace}.

After some related work overview in Section 2, QA generation and filtration details are demonstrated in Sections 3 and 4 respectively, followed by the corpus statistics in Section 5. Evaluation details are described in Sections 6 and 7. 

\section{Related Work}

The proposed work is based upon the recent architectures in transformer language modelling - GPT-3~\citep{brown2020language} and T5~\citep{raffel2019exploring}, and solves a standard SQuAD format problem, resulting in triplets \textit{"text paragraph - question based on paragraph - answer from the paragraph"}, see the following example: \begin{itemize}
    \item \textbf{Original Wikipedia paragraph:\footnote{\url{https://en.wikipedia.org/wiki/K\%C5\%8Dichi\_Mashimo}}} \begin{otherlanguage}{russian}\textit{Коити Масимо (яп.  Масимо Ко:ити) — известный режиссёр аниме и основатель японской анимационной студии Bee Train. С момента основания студии он руководит производством почти всех её картин, а также время от времени принимает участие в работе над анимацией и музыкой. }\end{otherlanguage}
    \textit{Kōichi Mashimo is a famous anime director and the founder of the Japanese animation studio Bee Train. Since the creation of the studio, he directed almost all studio's works, and he also sometimes participates in art and sound tasks.}
    \textbf{Generated question (ruT5): } \begin{otherlanguage}{russian}\textit{Кто является основателем японской анимационной студии Bee Train? } \end{otherlanguage}
    \textbf{Generated answer (ruT5):}\begin{otherlanguage}{russian} \textit{ Коити Масимо }\end{otherlanguage}
    \textbf{English QA translation:} \textit{Who is the founder of the Japanese animation studio Bee Train? Kōichi Mashimo}
\end{itemize}

The following subsections of this section will break down previous work on the topic of QA datasets and their generation.

\textbf{Datasets.} For English, SQuAD 1.1~\citep{rajpurkar2016squad} consists of 107,785 question-answer pairs. SQuAD 2.0, combines SQuAD 1.1 questions with over 50,000 unanswerable questions (questions that cannot be answered based on the corresponding paragraph)~\citep{rajpurkar-etal-2018-know}. The following datasets for English were of comparable size or bigger. Trivia QA~\citep{joshi2017triviaqa} includes 95 thousand QA pairs. Natural Questions (NQ)~\citep{kwiatkowski2019natural} contains questions from Google search queries and corresponding spans from Wikipedia articles as answers: 307,373 training examples, 7,830 development and 7,842 test examples. With the development of deep learning models, over 80 new datasets on QA and reading comprehension appeared in the past two years~\citep{rogersgardner2021}. Several multilingual QA datasets contain Russian examples: MKQA~\citep{MKQA}, TYDI QA~\citep{clark-etal-2020-tydi}, a dataset for 7 languages~\citep{asai2020xor}.~\citet{artetxe-etal-2020-cross} conducted experiments on the Cross-lingual Question Answering Dataset (XQuAD) benchmark that consists of a subset from SQuAD v1.1 and its translations into 10 languages. 

Wikipedia is commonly used as a relevant source for new datasets: for example, \citet{yang-etal-2015-wikiqa} presented WIKIQA dataset of QA pairs. It contains 3,047 questions from Bing query logs, where each one is associated with a Wikipedia page. Manual annotation was used to check if a sentence from a page summary paragraph is the correct answer to the question. \citet{lewis-etal-2021-paq} automatically generated 65M QA pairs from Wikipedia paragraphs, using four steps with separate models: passage selection, possible answer extraction (with BERT), question generation (with BART), and filtering.

For Russian, SberQuAD~\citep{efimov2020sberquad}\footnote{\url{https://huggingface.co/datasets/sberquad}} is the main resource for the QA system development and evaluation. The dataset was created following the methodology of the original English SQuAD, it contains about 50 thousand QA pairs. No bigger QA datasets for Russian were created yet, and synthetic QA generation approaches were not applied to Russian yet. Although, pre-trained language models, which are suitable for generative tasks, might help create better QA systems: ruGPT-3 models (ruGPT3XL, ruGPT3Large, ruGPT3Medium, ruGPT3Small) and ruT5 models (ruT5-base, ruT5-large) exist for Russian and can be implemented for the task.

\textbf{Question-answer generation.}. Classical QA pair generation pipeline lets firstly choose among text points that should be asked, then ask a question based on these points, and after that find the most likely candidate from the answer spans in text~\citep{reddy-etal-2017-generating,du-etal-2017-learning,alberti-etal-2019-synthetic,lee-etal-2020-generating}. Joint models, for question and answer generation can be also used~\citep{shakeri2020,DBLPCui2021} - i.e. based on BART.~\citet{lyu2021improving} proposed BERT-based model which generates questions heuristically from summaries. Some filtering steps can be done after creating QA  too~\citep{alberti-etal-2019-synthetic,puri-etal-2020-training,lewis-etal-2021-paq}.~\citet{shakeri2020} proposed likelihood of the generated question-answers as a measure for it.

In the recent years, pre-trained language models as unsupervised open-domain QA systems, that incorporate factual knowledge, were studied~\citep{petroni-etal-2019-language,10.1162tacl_a_00324,jiang-etal-2020-x,kassner-schutze-2020-bert,Bouraoui_Camacho-Collados_Schockaert_2020} and criticized~\citep{cao-etal-2021-knowledgeable}. Other pre-trained language models were also examined for the task:~\citet{wang-etal-2021-generative} fine-tuned BART to answer closed-book questions, and~\citet{WangLiu2020} studied GPT-2-based models performance for constructing knowledge graphs. 

To the best of our knowledge, the only approach to GPT-based QA generation and filtration was suggested in~\citep{Bang2020}, who used a QA generation pipeline to generate diverse question-answer pairs from unlabeled text corpus. For question generation, GPT-2 small model, fine-tuned on SQuAD 1.1 training dataset, was used. To filter out low-quality generated data,  fine-tuned BERT-based QA model utilizing the SQuAD 1.1 dataset was used: examples were kept if F1 similarity score between the answer span and the answer span predicted by BERT-based QA was above 0.9. The performance of question generation was evaluated by BLEU, ROUGE-L, METEOR metrics. However, the approach was examined only for English.


\section{Implementation Details}
We used the biggest freely available Russian GPT3 model: ruGPT-3 XL. The model was trained using Deepspeed and Megatron code and had sparse attention blocks. Maximal sequence length for generation was 2048 tokens.
We fine-tuned the model on SberQuAD dataset with the following parameters: batch-size = 2, sequence length = 2048, learning rate = 0.000015. The model fine-tuning required 10 GPUs per worker, and it took 135,000 iterations. After that we ran parallel QA generation with the parameters: maximal length = 1048, beam search with 7 as a number of beams, all 3grams can only occur once, repetition penalty = 2. 

We also fine-tuned ruT5-large model for Russian on SberQuAD dataset with such parameters: number of epochs = 5, maximal length = 512, batch size = 16, number of beams = 12. 

For ruGPT-3 XL, we turned each example into a line starting with a special text beginning token ($<$[TEXT]$>$), a text, then a special question beginning token ($<$[QUESTION]$>$), a question, a special answer beginning token ($<$[ANSWER]$>$), and, finally, an answer, followed with the end-of-sequence special token. For ruT5-large, we presented each example in the same way, but special text beginning, question beginning and answer beginning tokens were in Russian. Both models were fine-tuned to generate 3 QA pairs for a text. 

For QA generation we crawled all Wikipedia for the Russian language (up to March 2021) - 2,682,680 articles in general. We took only text from summary sections in every Wikipedia article. Based on page categories, we excluded disambiguation articles from the data. Then we kept Wikipedia article categories for each summary, for filtration and analysis purposes. For processing purposes, we splitted all Wikipedia data into 20 batches. Both for ruGPT-3 XL and for ruT5-large, we generated 3 QA pairs per summary. So the dataset contains summaries, QA pairs for them, and additional information, such as page title and corresponding Wikipedia categories. All QA pairs for a summary are included in one batch, and each summary appeared in the Wikipedia summaries dataset only once.

The dataset is presented in 20 batches, it lets use any 18 batches as train set, and the remaining two batches as development set and test set, if needed. 
Both ruGPT-3 XL and ruT5-large fine-tuning, generation, filtration and evaluation tasks were performed on 4 Tesla V100 GPU (32GB RAM) server and in Google Colab.


\section{Filtration of Generated Data}
Inspired by~\citep{Bang2020}, we applied a set of hand-crafted heuristics to filter out generated QA pairs of poor quality in the following steps, based on manual evaluation (See Subsection 6.1.). 
\begin{enumerate}
    \item First of all, we dropped QA pairs with more than one  interrogative pronoun in a question. 
    \item Then we applied squad\_ru\_rubert\_infer BERT model for Russian pre-trained on SberQuad \footnote{\url{http://docs.deeppavlov.ai/en/master/features/models/squad.html}}. We created 'gold' answers for all generated questions with it, letting it answer the questions generated by ruGPT-3 or ruT5. After that we left strings with exact match between lemmatized generated answer and BERT model answer, with intersection of lemmas between two answers over 70\%. This threshold was chosen manually based on the analysis of one data sample - batch 2 (random 50,000 examples from 90,927 summaries). 
    \item After that, we extracted named entities using Natasha python library for Russian.\footnote{\url{https://github.com/natasha/natasha}} We removed QA pairs in which entities (of different types) in a generated question were not presented in Wikipedia summary, and/or entities (of different types) in a generated answer were not presented in summary, using string match methods. 
    \item Finally, we deleted duplicated QA pairs for the same summaries where Levenshtein distance similarity ratio between questions and Levenshtein distance similarity ratio between answers was more than 70\%.
\end{enumerate}


Several additional options were implemented and can be used too, but they were not included into the final heuristics version for this specific task after the manual analysis (See Subsection 6.1.): 1)  ROUGE~\citep{lin-2004-rouge}, METEOR~\citep{banerjee-lavie-2005-meteor,denkowski-lavie-2014-meteor} and BLEU~\citep{10.3115/1073083.1073135} metrics in the second step. For each pair among \textit{'gold' answer - generated answer, question - generated answer, text - generated answer, text - question}, three metrics for lemmatized strings were calculated. The mean result for each pair was counted, and QA pairs where values were less than the corresponding thresholds 60\%, 50\%, 40\%, were removed. 2) Matching persons and locations separately instead of the third step. 3) Checking if the 'gold' BERT model score is over 0.99, filtering out complicated examples. 4) Calculating if word mover's distance between generated answer and 'gold' answer is between 1.1 and 1.5, using the fastText model for Russian\footnote{araneum fasttextcbow-300-5-2018.model \url{https://rusvectores.org/en/models/}}.

\begin{figure*}[!htp]
\centering
\includegraphics[width=13cm]{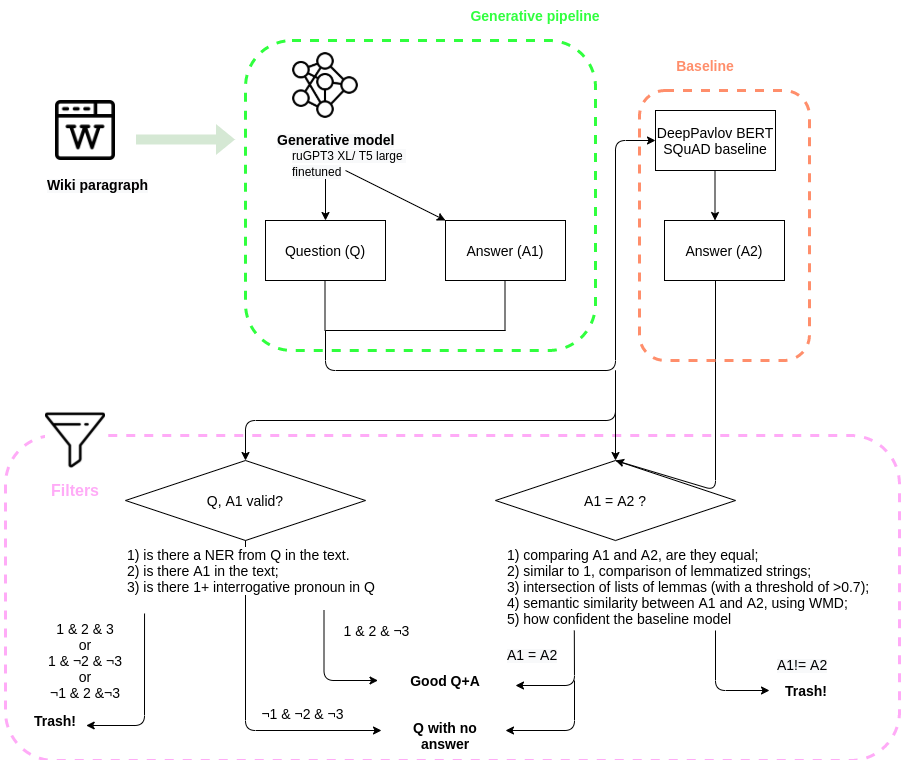}
\caption{The full WikiOmnia pipeline for QA generation.}
\label{fig:pipe}
\end{figure*}

The overall pipeline is presented in Figure~\ref{fig:pipe}.

\section{Corpus Statistics}
We describe the main characteristics of the resulting corpus. For synthetic data, it is especially important to control their diversity and frequency of words. 

\textbf{Basic statistics.} For ruGPT-3 and ruT5 generated data, generation and filtration results in a detailed way are presented in Tab.~\ref{tab:genfilt} for Wikipedia batches 1-5. We see that quality of ruT5 generated QA pairs is much better; however, both models require a filtration step for a 'clean' dataset version. In general, the raw dataset version for ruGPT-3 contained 7,930,873 examples, and filtered version had more than 160,000 examples. For ruT5, the raw dataset version consisted of 7,991,040 examples; filtered version included over 3,400,000 examples. 

The most frequent words in questions and answers for all 4 setups do not differ: they are about years, names, places, numbers etc. For instant, in questions the most frequent lemmas are 'god' (\textit{year}), 'nazyvat'sja' (\textit{to be named}), 'rodit'sja' (\textit{to be born}), 'skol'ko' (\textit{how many}), 'gorod' (\textit{town}), and in answers the most frequent lemmas are 'god' (\textit{year}), 'rajon' (\textit{district}), 'chelovek' (\textit{human, person}), 'gorod' (\textit{town}), 'rossijskij' (\textit{Russian}).\footnote{Here Russian words are given in Latin transliteration, for readability purpose.} Average length for ruT5 before and after filtration is about 52 characters (7 tokens) for questions and about 24 characters (4 tokens) for answers. For ruGPT-3, average length before filtration is 47 characters (7 tokens) for questions and 19 characters (3 tokens) for answers; after filtration its is slightly shorter: 46 characters (7 tokens) for questions and 12 characters (2 tokens) for answers. In SberQuAD train set, questions (64.4 characters, 8.7 tokens) and answers (25.9 characters, 3.7 tokens) are longer~\citep{efimov2020sberquad}.

\begin{table*}[!h]
\centering
\begin{tabular}{lllll}
\hline

\textbf{Batch}&\textbf{ruGPT-3 before filtering}&\textbf{filtered ruGPT-3}&\textbf{ruT5 before filtering}&\textbf{filtered ruT5}\\
\hline
Batch1 & 266,332& 10,079& 272,397& 152,884\\
Batch2 & 268,795& 8,034& 271,281& 113,964\\
Batch3 & 276,618& 6,176& 275,412& 124,784\\
Batch4 & 272,875& 7,042& 270,534& 146,627\\
Batch5 & 276,107& 5,536& 279,363 &157,535\\
\hline

\end{tabular}
\caption{Number of QA pairs in ruGPT-3 and ruT5 generated batches before and after filtering: on the example of Wikipedia batches 1-5.}
\label{tab:genfilt}
\end{table*}

\textbf{Self-BLEU for questions diversity.} We computed Self-BLEU as a metric of diversity for generated questions, as they are more specific for a model than answers that depend on questions. We followed~\citep{holtzman-etal-2020-curious} approach that is based on~\citep{Zhu2018TexygenAB}. It yields how one sentence (a question) resembles other generated questions in the collection: for each question as a hypothesis and all other questions as references, the BLEU score is calculated. Due to computational reasons, we took random samples of 5,000 examples from raw ruGPT-3 data (batch 2), raw ruT5 data (batch 2), filtered ruGPT-3 data (including batch 2), and filtered ruT5 data (including batch 2). To compare, we measured Self-BLEU for a random sample of 5,000 questions from the original SberQuAD too. For each text, there was only one corresponding question in the data. Questions were lemmatized before calculation.

Median Self-BLEU scores are presented in Tab.~\ref{tab:selfbleu}, where lower Self-BLEU scores represent higher diversity. SberQuAD data demonstrated the  highest diversity. While ruT5 generated questions imply higher diversity  after filtration, for ruGPT-3 the most relevant questions that remain after filtration are less diverse. 

\begin{table}[!h]
\centering
\begin{tabular}{lc}
\hline

\textbf{Data}&\textbf{Median Self-BLEU} \\
\hline
Raw ruGPT-3 data (1) & 0.45\\
Filtered ruGPT-3 data (2) & 0.49\\
Raw ruT5 data (3) & 0.40\\
Filtered ruT5 data (4) & 0.38\\
SberQuAD data (5) & 0.20\\
\hline

\end{tabular}
\caption{Median Self-BLEU scores calculated for raw ruGPT-3 generated data (1), filtered ruGPT-3 generated data (2), raw ruT5 generated data (3), filtered ruT5 generated data (4), SberQuAD data (5).}
\label{tab:selfbleu}
\end{table}

\begin{table*}[!h]
\centering
\begin{tabular}{lccccc}


\hline
\textbf{Wh-word}&\textbf{1}&\textbf{2}&\textbf{3}&\textbf{4}&\textbf{5}\\
\hline
kto (\textit{who}) & 0.15 & 0.04& 0.14& 0.12&0.05\\
chto (\textit{what}) & 0.08 & 0.10& 0.06& 0.07&0.13\\
kakoj (\textit{which, what}) & 0.08 & 0.12& 0.09& 0.08&0.11\\
gde (\textit{where}) & 0.07 & 0.04& 0.12& 0.10&0.03\\
skol'ko (\textit{how many}) & 0.02 & 0.06& 0.04& 0.04&0.03\\
kakov (\textit{what, which}) & 0.05 & 0.02& 0.01& 0.00&0.01\\
kogda (\textit{when}) & 0.06 & 0.04& 0.11& 0.12&0.05\\
pochemu (\textit{why}) & 0.00 & 0.00& 0.00& 0.00&0.01\\
chem (\textit{with what}) & 0.01 & 0.01& 0.01& 0.01&0.03\\
kak (\textit{how}) & 0.18 & 0.13& 0.14& 0.11&0.07\\
\hline
\end{tabular}
\caption{Wh-questions median ratios for raw ruGPT-3 generated data (1), filtered ruGPT-3 generated data (2), raw ruT5 generated data (3), filtered ruT5 generated data (4) on the example of Wikipedia batches 1-5; Wh-questions ratio for SberQuAD data (5).}
\label{tab:wh}
\end{table*}

\textbf{Wh-questions ratio.} We also use wh-questions ratio to check how diverse are the questions. We select 15 Wh-words and similar words in Russian: 'kto' (\textit{who}), 'chto' (\textit{what}), 'kakoj' (\textit{which, what}), 'chej' (\textit{whose}), 'gde' (\textit{where}), 'kotoryj' (\textit{what, which}), 'otkuda' (\textit{where from}), 'skol'ko' (\textit{how many}), 'kakovoj' (\textit{what, by which}), 'kakov' (\textit{what, which}), 'zachem' (\textit{what for}), 'kogda' (\textit{when}), 'pochemu' (\textit{why}), 'chem' (\textit{with what}), 'kak' (\textit{how}).\footnote{Here Russian words are given in Latin transliteration, for readability purpose.} On the example of 5 batches, we checked how many such questions were presented in data before and after filtration, compared with SberQuAD ratios. Tab.~\ref{tab:wh} demonstrates results for 10 Wh-words and similar words, excluding 'chej' (\textit{whose}), 'otkuda' (\textit{where from}), 'zachem' (\textit{what for}), 'kotoryj' (\textit{what, which}),  'kakovoj' (\textit{what, 'by which'}), that were underrepresented both in 5 batches and in SberQuAD. For both ruGPT-3 and ruT5 generated questions, ratios for 'skol'ko' (\textit{how many}) and 'kak' (\textit{how}) after filtration are higher than in SberQuAD questions. Generated QA pairs of good quality more often contain a numerical answer. Questions with 'kto' (\textit{who}), 'gde' (\textit{where}), and 'kogda' (\textit{when}) have higher ratios in ruT5 questions than in SberQuAD. On the contrary, more complicated questions with 'pochemu' (\textit{why}) and 'chem' (\textit{with what}) are less presented in generated QA pairs. It can be also noticed that ruGPT-3 generates questions with 'kakov' (\textit{what, which}) (a short form of a wh-word) more often than ruT5. ruGPT-3 generated QA pairs with 'kto' (\textit{who}) have rather low quality and contain information about persons not from the summaries, that's why they are strictly filtered out. On the example of 'kogda' (\textit{when}), we see that QA pairs with dates, provided by ruT5, are more correct than such pairs from ruGPT-3. Therefore, in comparison with SberQuAD, both generated datasets remain diverse, too.

%
 %


\section{Performance Evaluation}

\subsection{Human Evaluation and Error Analysis}

\textbf{Human Evaluation for editing the pipeline.} We took human evaluation into account for the data generated by a fully automated generative pipeline twice, conducting the intermediate and the final evaluation stages. This manual evaluation was conducted by the authors, as well as discussions about problematic points to handle disagreements. On the intermediate stage, we took multiple series of 10,000 random summaries and analysed manually the generated QA pairs for them, as well as the examples remaining after filtration with different filtration options; the same steps were reproduced for QA pairs by ruGPT-3 and ruT5. Based on this intermediate evaluation, the generation and filtration pipeline was edited: step 1 was added; step 3 was placed after step 2 (not before it); several steps were removed from the pipeline (See Section 4). After that, the final evaluation stage was conducted for the same samples with the final filtration options results: for these samples, rate of examples remaining after filtration reached about 5\% for QA pairs generated by ruGPT-3 and about 30\% for QA pairs generated by ruT5. During the final stage, we also checked manually, in addition, several random samples of 10,000 QA pairs for specific evaluation tasks.   

\textbf{Wikipedia topics before and after filtration.} To estimate if filtration ratio varies for different topics, we checked the ratio of examples that remained after filtration for various Wikipedia categories groups (on the example of Batches 1-5): history events, famous persons biographies, plants, technical descriptions, geography, mathematics, sports, actors, and movies. Categories for the selected topics were grouped using heuristics rules, based on saved Wikipedia category names for each example (one example could have multiple categories).

Both ruGPT-3 and ruT5 generated QA pairs showed the best results for articles about sports, perhaps due to simple and well-structured summaries. Error analysis showed that ruGPT-3 also performed rather well on history and plants topics, but answers to the correct questions, also correct in meaning, did not match the 'gold' answers well. In addition to sports, ruT5 QA pairs for technical, history and geography articles also yielded higher quality, they did not contain additional information not from the corresponding summaries, unlike ruGPT-3. In general, for technical topics (i.e. computer science), generated QA pairs yielded worse quality before filtration than for other topics.

Example of an erroneous QA pair generation with ruT5, detected by filtering:
\begin{itemize}
    \item \textbf{Original Wikipedia paragraph:\footnote{\url{https://en.wikipedia.org/wiki/Psathyrella\_piluliformis}}} \begin{otherlanguage}{russian}\textit{Псатирелла водолюбивая (лат. Psathyrella piluliformis) — гриб рода Псатирелла (Psathyrella) семейства Псатирелловые (Psathyrellaceae). Съедобность гриба спорна, чаще он считается несъедобным, иногда — условно съедобным, но невысокого качества. }\end{otherlanguage} 
    \textit{Psathyrella piluliformis is a species of agaric fungus in the family Psathyrellaceae. It is considered edible but of low quality, with fragile flesh and being difficult to identify.}
    \textbf{Generated question (ruT5):}  \begin{otherlanguage}{russian}\textit{Какова способность гриба менять окраску?} \end{otherlanguage}
    \textbf{Generated answer (ruT5): }\begin{otherlanguage}{russian} \textit{в зависимости от условий }\end{otherlanguage} 
    \textbf{English QA translation:} \textit{What is the ability of a fungus to change color? Depends on conditions}
\end{itemize}

The filtered dataset may still contain two types of errors that were not detected by the filters: 1) questions about information that was not presented in a summary (0.008\% for ruGPT-3, based on a random example of 10,000 QA pairs); 2) erroneous answers with numbers (if not years). 

\subsection{Automated Evaluation}

During all evaluation experiments, we focused mostly on training QA systems using the filtered WikiOmnia part with QA pairs generated by ruT5, as it is bigger (than the part with QA pairs by ruGPT-3) and lets experiment with different sample sizes. We took random dataset parts of 50,000 examples, 100,000 examples, and 300,000 examples. For each sample size, we took 2 random samples and calculated the average score values for them.

\textbf{Experiment set 1.} We fine-tuned ruBERT base cased model (BERT model for Russian \footnote{\url{https://huggingface.co/DeepPavlov/rubert-base-cased}}) on each of these samples and then evaluated it on development and test parts of SberQuAD dataset. As a baseline, we fine-tuned ruBERT on the train part of SberQuAD dataset. F1 score and exact match (EM) were used as standard SQuAD evaluation metrics. For all setups, the following parameters were used for fine-tuning: 3 epochs, learning rate = 2e-5, weight decay = 0.01. 

\textbf{Experiment set 2.} We took models above, already fine-tuned on WikiOmnia samples (100,000 examples and 300,000 examples), and fine-tuned them further on SberQuAD train part (1, 2 and 3 epochs). Results for  Experiment sets 1 and 2 are presented in Tab.~\ref{tab:finetuning}. In the second experiment set, the models fine-tuned on 100,000 or 300,000 WikiOmnia triplets and then fine-tuned on SberQuAD train part (2 epochs), perform better than models fine-tuned only on 100,000 or 300,000 WikiOmnia triplets, or the baseline model fine-tuned on SberQuAD train part (3 epochs).  Fine-tuning first on WikiOmnia and then on SberQuAD yields better results than fine-tuning only on SberQuAD.\footnote{Models, fine-tuned on the ruGPT-3 generated WikiOmnia part, showed the same peruliarity: after fine-tuning on WikiOmnia and then on SberQuAD train, EM on the development set was 67.71, F1 score on the development set was 86.64, EM on the test set was 66.57, and F1 score on the test set was 85.88. All metrics, excepting the last one, are better than the baseline. As the filtered ruGPT-3 generated WikiOmnia part is rather small and contains only 164,253 examples, all experiments were conducted for this whole part.} The WikiOmnia size lets conduct experiments with different sample sizes. 

\begin{table*}[!h]
\centering
\begin{tabular}{lcccc}

      \hline
      \textbf{Model setup}&\textbf{EM on dev}&\textbf{F1 on dev}&\textbf{EM on test}&\textbf{F1 on test}\\
      \hline
      Baseline (1) & 66.39& 85.92& 66.46& 85.93\\
      (2) & 59.26& 79.89& 58.30& 79.26\\
      (3) & 60.04& 80.57& 58.64& 79.94\\
      (4) & 59.80& 80.50& 58.36& 79.97\\
      (5) & 67.32& 86.26& 66.29 &85.76\\
      (6) & 67.04& 86.18& 66.96 &86.03\\
      (7) & 65.95& 85.67& 65.48 &85.40\\
      \hline

\end{tabular}
\caption{Evaluation scores for ruBERT base model fine-tuned on: SberQuAD train (1), WikiOmnia 50,000 examples by ruT5 (2); WikiOmnia 100,000 examples by ruT5 (3); WikiOmnia 300,000 examples by ruT5 (4); WikiOmnia 100,000 examples and then SberQuAD train 1 epoch (5); WikiOmnia 100,000 examples and then SberQuAD train 2 epochs (6); WikiOmnia 100,000 examples and then SberQuAD train 3 epochs (7).}
\label{tab:finetuning}
\end{table*}

\textbf{Experiment set 3.} Following the Experiment set 2 results, we decided to take an 'own' development set from WikiOmnia (10,000 triplets) and to compare results on it with results on development and test parts of SberQuAD (Tab.~\ref{tab:evalcompwo}). We took a random sample with 110,000 examples from WikiOmnia by ruT5. We conducted ruBERT base model fine-tuning: 5 runs for different folds where 10,000 triplets were taken as a development set for evaluation, and the remaining 100,000 triplets were used for fine-tuning, 2 epochs in each run. Results on the WikiOmnia development set, in all runs, are much better than results on SberQuAD development and test sets. Perhaps, due to the datasets specifics, SberQuAD development and test sets are suitable for models, fine-tuned on WikiOmnia, evaluation, only if they were fine-tuned on SberQuAD train as a second step. 

\begin{table*}[!h]
\centering
\begin{tabular}{lcccccc}

      \hline
      \textbf{Model}&\textbf{EM on own dev}&\textbf{F1 on own dev}&\textbf{EM on dev}&\textbf{F1 on dev}&\textbf{EM on test}&\textbf{F1 on test}\\
      \hline
      1 run & 87.47& 95.14& 59.32& 80.00& 58.30& 79.71\\
      2 run & 87.89& 95.24& 59.86& 80.40& 58.46& 79.87\\
      3 run & 88.08& 95.46& 59.83& 80.33& 58.57& 79.92\\
      4 run & 87.53& 95.18& 60.22& 80.66& 58.64& 79.89\\
      5 run & 87.90& 95.18& 59.39 &80.18& 58.21& 79.70\\
      \hline
\end{tabular}
\caption{Evaluation on the development set from WikiOmnia (own dev), in comparison with evaluation on SberQuAD development (dev) and test (test) sets (5 runs).}
\label{tab:evalcompwo}
\end{table*}

\section{Pipeline Evaluation on Other Domains}

For evaluation purposes, we also tested the full pipeline on data samples of four other text genres in Russian: news stories, social media posts, product reviews, and fiction texts. Each sample has 2,000 examples taken randomly from the following datasets: 1) news from the newspaper  Gazeta,\footnote{\url{https://github.com/IlyaGusev/gazeta}}~\citep{Gusev2020gazeta} with text lengths up to 3,500 characters; 2) social media texts from the Taiga corpus\footnote{\url{https://tatianashavrina.github.io/taiga\_site/}} ~\citep{shavrina2017}, with texts lengths up to 3,000 characters;  3) reviews from the dataset\footnote{\url{https://github.com/sismetanin/rureviews}} on product reviews about clothes from an e-commerce website~\citep{Smetanin-SA-2019}, with text lengths over 500 characters and up to 1,007 characters, as these texts are rather short; 4) fiction texts from the collection of Russian classical literature texts\footnote{\url{https://www.kaggle.com/d0rj3228/russian-literature}}: fragments from texts, with text lengths up to 3,000 characters.

For every text, three QA pairs were generated. Filtration steps were the same as for QA pairs based on Wikipedia summaries. After filtration, we got the following results for ruGPT3XL: 497 pairs remained for fiction texts, and 559 pairs were left for news texts. For reviews, the pipeline performed in the best way: 1379 pairs were left. The worst performance was for social media: only 154 pairs remained. ruT5-large also yielded good performance on reviews: 1,542 pairs were left after filtration. The explanation might be that review texts as a genre usually have definite patterns and structure. The worst ruT5-large results were also for social media: only 945 pairs remained. Social media texts looked mostly like opinionated pieces where it would be hard to create QA pairs manually too. 

Both pipelines, for ruGPT3 XL and ruT5-large, can be generalized comparatively well to other genres. Although ruT5-large performed generally better on all four genres, the results mostly differed on news texts: 3,204 texts remained after filtering. 
Other filtration techniques should be investigated, handling the remaining errors, i.e. how to check quality of numerical answers (especially by ruGPT-3), or how to check question and answer similarity to the corresponding summary, considering paraphrases. Reasons of the results of the automated evaluation on SberQuAD development and test sets should be also explored further. The dataset implementation for various and diverse tasks and its evaluation on them remains a separate point for further research. 

\section{Conclusions}

We propose WikiOmnia, the new largest question-answering dataset for  Russian: it contains QA pairs and corresponding Russian Wikipedia article summaries. It can be used to improve the quality of monolingual and multilingual information retrieval systems, open domain question answering, etc. Quality of generated QA pairs in the filtered part of the dataset is ensured by diverse automated filtration techniques, manual and automated evaluation.  
We also present the automated generation and filtration pipeline that can be applied to various sources of text data, including expanding the applicability of QA systems to news data, fiction, reviews. 

We welcome researchers in the fields of information retrieval and language technology to use both the dataset to train the models, and the pipeline to expand the capabilities and robustness of the existing QA systems. We invite the community to reproduce the work on materials of other languages, using multilingual models and existing baselines.


\bibliography{anthology,custom}
\bibliographystyle{acl_natbib}

\end{document}